\documentclass{article}

\usepackage[cmex10]{amsmath}
\usepackage{amssymb}
\usepackage{graphicx}
\usepackage{color}
\usepackage{array}
\usepackage{multirow}

\usepackage{algorithm}
\usepackage{algorithmic}

\usepackage{url}
\newcommand{\vc}{\mathbf{c}}

\newcommand{\vs}{\mathbf{s}}

\newcommand{\vu}{\mathbf{u}}

\newcommand{\vx}{\mathbf{x}}
\newcommand{\vy}{\mathbf{y}}

\newcommand{\vX}{\mathbf{X}}
\newcommand{\vY}{\mathbf{Y}}

\begin{document}

\title{MMSE Estimation for Poisson Noise Removal in Images}

\author{Stanislav~Pyatykh* and~J{\"u}rgen~Hesser
\thanks{The authors are with the University Medical Center Mannheim, Heidelberg University,
Theodor-Kutzer-Ufer 1-3, D-68167 Mannheim, Germany
(e-mail: stanislav.pyatykh@medma.uni-heidelberg.de; juergen.hesser@medma.uni-heidelberg.de).}}

\maketitle

\begin{abstract}
Poisson noise suppression is an important preprocessing step in several applications, such as medical imaging, microscopy, and astronomical imaging.
In this work, we propose a novel patch-wise Poisson noise removal strategy,
in which the MMSE estimator is utilized in order to produce the denoising result for each image patch.
Fast and accurate computation of the MMSE estimator is carried out using k-d tree search followed by search in the K-nearest neighbor graph.
Our experiments show that the proposed method is the preferable choice for low signal-to-noise ratios.
\end{abstract}

%\begin{IEEEkeywords}
%Estimation, image processing, image denoising.
%end{IEEEkeywords}

%\IEEEpeerreviewmaketitle

\section{Introduction}
\label{introduction}

Images corrupted with Poisson noise appear in many applications, such as medical imaging, fluorescence microscopy, and astronomical imaging.
There, the noise removal quality plays an essential role for the further image processing steps.
For Poisson noise, the noise variance equals the true image intensity,
which means that noise is signal-dependent and the signal-to-noise ratio equals the square root of the image intensity.
For this reason, the methods developed for additive white Gaussian noise are not applicable for Poisson noise directly.

Many of currently existing approaches reduce the Poisson noise removal to the Gaussian noise removal.
This is an attractive strategy since the state of the art algorithms for Gaussian noise suppression are very efficient:
it was shown in \cite{levin2011natural,levin2012patch} that they have almost achieved theoretical performance boundary.
The reduction of the denoising problem for Poisson noise to that for Gaussian noise is based on the variance stabilization \cite{boulanger2010patch,makitalo2011optimal,zhang2008wavelets}.
First, a pixel-wise variance-stabilizing transformation (such as the Anscombe root transformation \cite{anscombe1948transformation}) is applied in order to
transform the input image with Poisson noise into the image with signal-independent Gaussian noise.
Second, a method for Gaussian noise removal is utilized (e.g. \cite{dabov2007image,elad2006image,burger2013learning}).
Finally, an inverse transformation is applied.
In \cite{makitalo2011optimal} and \cite{foi2009clipped}, it was pointed out that application of the algebraic inverse introduces a bias and the exact unbiased inverse was developed.

Nevertheless, the variance stabilization becomes inaccurate as the image intensity decreases,
which limits the applicability of the above mentioned approaches.
This fact led to development of denoising methods, which work with Poisson-distributed data directly.
Among them, \cite{salmon2014poisson} and \cite{giryes2014sparsity} are the most efficient.
First, these algorithms group similar image patches into clusters.
Next, patches of each cluster are represented using a small number of dictionary elements.
Due to the projection onto the subspace defined by these dictionary elements, the noise components orthogonal to this subspace are canceled out.
Therefore, the dimensionality of this subspace, i.e. the number of the dictionary elements used to represent each patch, should be made as small as possible.
In \cite{salmon2014poisson}, the dictionary elements are computed using Poisson principal component analysis of each patch cluster,
which is based on the Bregman divergence for Poisson distribution.
In \cite{giryes2014sparsity}, the dictionary elements are selected using a greedy algorithm,
which searches for the maximum of the likelihood of the denoised patch given the input noisy patch.

In this paper, we propose a different technique for Poisson noise removal.
Instead of projecting image patches onto a subspace defined by a dictionary,
the denoised patch is computed as the minimum mean square error (MMSE) estimator of the true patch given the noisy patch.
If all possible noise-free image patches were available for the algorithm as prior information,
such approach would provide the optimal estimate of the true patch.
However, we show that it outperforms existing methods also with computationally tractable amount of prior information.

The MMSE estimation was considered in the image denoising context in \cite{levin2011natural,levin2012patch},
where it was used to compute the theoretical performance bounds of denoising algorithms.
Since the direct computation of the MMSE estimator is very time-consuming, practical algorithms based on the MMSE estimation were not developed in these works.
An optimized denoising algorithm based on the MMSE estimation was presented in \cite{lee2012mmse} for the case of Gaussian noise,
however, its speed drops as signal-to-noise ratio decreases.
In this paper, we propose a fast and efficient method to compute the MMSE estimator for an image patch.
This leads to a practical image denoising algorithm, which is able to process images in a reasonable time and outperforms existing approaches in terms of the output image quality.

\section{MMSE estimation for Image Patches}
\label{MMSE_estimation}

Consider an unknown noise-free image patch of size $\sqrt{d} \times \sqrt{d}$ rearranged into vector $\vx \in \mathbb{R}^d$
and the corresponding patch of the input noisy image represented as vector $\vy \in (\mathbb{N} \cup 0)^d$.
We model patch $\vx$ using random vector $\vX \in \mathbb{R}^d$ with probability density function $f_\vX(\cdot)$,
whereas patch $\vy$ is modeled by random vector $\vY \in (\mathbb{N} \cup 0)^d$ with probability mass function $f_\vY(\cdot)$.
Assume that we know the likelihood $P(\vY_i=y|\vX_i=x)$, where $\vX_i$ and $\vY_i$ are the $i$-th elements of $\vX$ and $\vY$ respectively.
In the case of Poisson noise,
\begin{align}
\label{Poisson_likelihood}
P(\vY_i=y|\vX_i=x) =
\begin{cases}
x^y e^{-x} / y! & x > 0 \\
0 & x = 0, y > 0 \\
1 & x = 0, y = 0
\end{cases}
\end{align}

Let us denote the peak intensity of the noise-free image by $I_{max}$.
The problem to denoise patch $\vy$ is typically formulated in terms of the peak signal-to-noise ratio (PSNR):
we look for estimate $\hat{\vx} \in \mathbb{R}^d$, which maximizes 
\begin{align}
\text{PSNR} = 10 \log_{10} \frac{I_{max}^2}{\text{MSE}}
\end{align}
or, equivalently, which minimizes the mean square error (MSE):
\begin{align}
\text{MSE} & = E( \lVert \hat{\vx} - \vX \rVert^2 | \vY = \vy ) \nonumber \\
& = \int \lVert \hat{\vx} - \vu \rVert^2 f_{\vX}(\vu|\vY=\vy) d\vu.
\end{align}
Such $\hat{\vx}$ is the MMSE estimator \cite{levin2011natural}, which is computed as
\begin{align}
\label{cond_exp}
\hat{\vx} & = E( \vX | \vY = \vy ) = \int \vu f_{\vX}(\vu|\vY=\vy) d\vu.
\end{align}
Using Bayes' theorem and the properties of marginal distributions,
the latter expression can be rewritten as
\begin{align}
\label{MMSE_estimator}
\hat{\vx} & = \frac{1}{f_\vY(\vy)} \int \vu f_{\vY}(\vy|\vX=\vu) f_\vX(\vu) d\vu  \nonumber \\
& = \frac{\int \vu f_{\vY}(\vy|\vX=\vu) f_\vX(\vu) d\vu}{\int f_{\vY}(\vy|\vX=\vu) f_\vX(\vu) d\vu}
\end{align}
where $f_{\vY}(\vy|\vX=\vu)$ is the probability of $\vy$ given $\vu$, which can be computed using (\ref{Poisson_likelihood}):
\begin{align}
f_{\vY}(\vy|\vX=\vu) & = \prod_{i=1}^d P(\vY_i = \vy_i | \vX_i = \vu_i ).
\end{align}

Since the probability density function for image patches $f_\vX(\cdot)$ cannot be accurately represented in some analytic form,
direct computation of the integrals in (\ref{MMSE_estimator}) is not possible in practice.
In order to approximate these integrals, one can use a large number of natural image patches $\vu^{(1)}, \ldots, \vu^{(N_P)}$,
which are treated as realizations of random vector $\vX$ \cite{levin2011natural}:
\begin{align}
\label{MMSE_estimator_app}
\hat{\vx}_\text{app} = \frac{\sum_{k=1}^{N_P} \vu^{(k)} f_{\vY}(\vy|\vX=\vu^{(k)})}{\sum_{k=1}^{N_P} f_{\vY}(\vy|\vX=\vu^{(k)})}.
\end{align}
This expression gives a computationally feasible approximation for the MMMSE estimator, which can be utilized for denoising of patch $\vy$.

\section{Efficient Patch Denoising}

In order to approximate $\hat{\vx}$ by $\hat{\vx}_\text{app}$ accurately enough, one has to use about $N_P = 10^8$ natural image patches.
Therefore, direct evaluation of the sums in (\ref{MMSE_estimator_app}) is very computationally expensive and, therefore, inapplicable.
In this section, we present a fast and practical method to compute $\hat{\vx}_\text{app}$.

\subsection{Offline step}

Several data structures are constructed based on patches $\vu^{(1)}, \ldots, \vu^{(N_P)}$ at the offline (pre-processing) step.

First, patches $\vu^{(1)}, \ldots, \vu^{(N_P)}$ are normalized
since the intensity range of images from which they are extracted does not correspond to the intensity range of the input image.
Specifically, each patch $\vu^{(k)}$ is divided by the average of all intensities occurring in $\vu^{(1)}, \ldots, \vu^{(N_P)}$,
resulting in normalized patch $\overline{\vu}^{(k)}$:
\begin{align}
\overline{I} & = \frac{1}{d N_P}\sum_{k=1}^{N_P} \sum_{i=1}^d \vu_i^{(k)} \nonumber \\
\overline{\vu}^{(k)} & = \vu^{(k)} / \overline{I} \quad k = 1, \ldots, N_P.
\end{align}

Next, we apply the k-means clustering to patch set $\overline{\vu}^{(1)}, \ldots, \overline{\vu}^{(N_P)}$.
For each cluster $j$, it provides the cluster centroid $\vc^{(j)}$ and the number of patches assigned to this cluster $n_j$.
Since each patch $\overline{\vu}^{(k)}$ can be approximated by the centroid of the cluster it is assigned to,
(\ref{MMSE_estimator_app}) can be replaced with
\begin{align}
\label{MMSE_estimator_cluster}
\hat{\vx}_\text{app} = \frac{\sum_{j=1}^{N_C} \vc^{(j)} n_j f_{\vY}(\vy|\vX=\vc^{(j)})}{\sum_{j=1}^{N_C} n_j f_{\vY}(\vy|\vX=\vc^{(j)})}
\end{align}
where $N_C$ is the number of clusters.

Then, centroids $\vc^{(1)}, \ldots, \vc^{(N_C)}$ are used to construct a set of $N_T$ randomized k-d trees as described in \cite{muja2014scalable}.
For each node of a randomized k-d tree, the splitting dimension is chosen randomly from several dimensions with the largest variance,
so that each k-d tree represents a different space partitioning.

Additionally, the $K$-nearest neighbor graph is constructed using centroids $\vc^{(1)}, \ldots, \vc^{(N_C)}$ as nodes.
For each centroid $\vc^{(j)}$, we compute the neighbor list containing the indexes of $K$ centroids from $\vc^{(1)}, \ldots, \vc^{(N_C)}$,
which are the nearest to $\vc^{(j)}$ in the Euclidean metric (excluding $\vc^{(j)}$ itself).

\subsection{Online step}

\begin{algorithm}[H]

\caption{}
\begin{algorithmic}[1]
\STATE $\vs \leftarrow d$-dimensional zero vector
\STATE $w \leftarrow 0$
\STATE $\mu_\vy \leftarrow \frac{1}{d} \sum_{i=1}^d \vy_i$
\STATE $Q \leftarrow $ empty priority queue
\STATE Mark all clusters as unprocessed
\FORALL{ k-d tree $T$ }
    \STATE $\{ j_m \}_{m=1}^L  \leftarrow$ leaf of $T$ containing $\vy/\mu_\vy$
    \STATE ProcessClusters$\big(\{ j_m \}_{m=1}^L\big)$
\ENDFOR
\WHILE{ $Q$ is not empty \AND $w$ has not converged}
    \STATE Pop element $j^*$ with the highest priority from $Q$
    \STATE $\{ j_m \}_{m=1}^K  \leftarrow$ neighbors of $\vc^{(j^*)}$
    \STATE ProcessClusters$\big(\{ j_m \}_{m=1}^K\big)$
\ENDWHILE
\STATE $\hat{\vx}_\text{app} \leftarrow \vs / w$
\RETURN $\hat{\vx}_\text{app}$
\end{algorithmic}
\begin{algorithmic}[1]
\REQUIRE ProcessClusters$\big(\{ j_m \})$:
\FORALL{ unprocessed cluster $j_m$}
    \STATE $\vs \leftarrow \vs + n_{j_m} f_{\vY}(\vy|\vX=\mu_\vy\vc^{(j_m)}) \cdot \mu_\vy \vc^{(j_m)}$
    \STATE $w \leftarrow w + n_{j_m} f_{\vY}(\vy|\vX=\mu_\vy\vc^{(j_m)})$
    \STATE Push $j_m$ into $Q$ with priority $f_{\vY}(\vy|\vX=\mu_\vy\vc^{(j_m)})$ 
    \STATE Mark cluster $j_m$ as processed
\ENDFOR
\end{algorithmic}

\end{algorithm}

In order to compute estimator (\ref{MMSE_estimator_cluster}) efficiently,
we utilize the fact that patches $\vc^{(j)}$ similar to input patch $\vy$ have the largest likelihood $f_{\vY}(\vy|\vX=\vc^{(j)})$ and,
therefore, the largest contribution to the sums in (\ref{MMSE_estimator_cluster}).
On the other hand, likelihood $f_{\vY}(\vy|\vX=\vc^{(j)})$ is almost zero for patches $\vc^{(j)}$, which are significantly different from $\vy$,
so that these patches have almost no influence on the result.
Therefore, we start the calculation of (\ref{MMSE_estimator_cluster}) from patches $\vc^{(j)}$, which are the closest to $\vy$,
and continue by processing patches farther away from $\vy$ until convergence of the result.

This procedure is presented in Algorithm 1, which takes noisy patch $\vy$ as the input and computes denoised patch $\hat{\vx}_\text{app}$.
Vector $\vs \in \mathbb{R}^d$ represents the numerator of (\ref{MMSE_estimator_cluster}), whereas $w \in \mathbb{R}$ stores the denominator.
First, centroids in the neighborhood of $\vy$ are found by traversing each k-d tree (lines 6--9).
The traversal starts from the root and goes to either the left or the right child depending on the location of $\vy$ relative to the splitting plane until a leaf node is reached.
Centroids $\{ \vc^{(j_m)} \}_{m=1}^L$ assigned to this node are then used to update $\vs$ and $w$ and added to the priority queue.
Next, centroids farther away from $\vy$ are found using the $K$-nearest neighbor graph (lines 10--14).
At each iteration of the loop, the centroid with the largest likelihood is removed from the priority queue and its previously unprocessed neighbors are examined.
The loop repeats until the convergence criterion is met (line 10), namely,
until the relative change of $w$ during the last $M$ iterations is less than $10^{-12}$.

Note that centroids $\vc^{(j)}$ were constructed using normalized patches $\overline{\vu}^{(k)}$.
Therefore, noisy patch $\vy$ should be normalized in order to be utilized for the k-d tree search (line 7) and,
vise versa, centroids $\vc^{(j)}$ should be multiplied by the mean of $\vy$ to compute $\vs$, $w$, and the likelihood (see the sub-algorithm ProcessClusters).

In order to denoise the input image, we extract its overlapping patches in a sliding window manner.
Next, these patches are denoised independently from each other using Algorithm 1 and returned to their original positions.
Since they are overlapped, there are several estimates for each pixel, which are then averaged to produce the final denoised image.

\section{Experiments}

\subsection{Setup}

\begin{figure}
	\centering
	\includegraphics[width=0.13\textwidth]{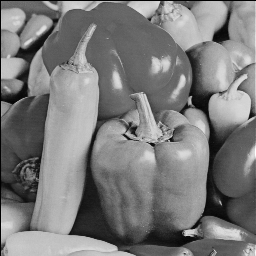}
	\includegraphics[width=0.13\textwidth]{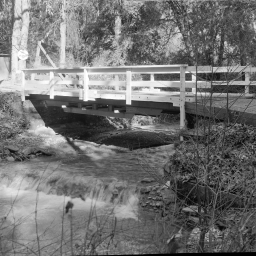}
	\includegraphics[width=0.13\textwidth]{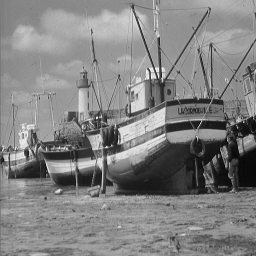}
	\includegraphics[width=0.13\textwidth]{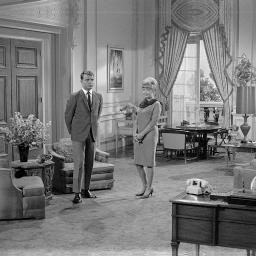}
%	\linebreak
%	\linebreak
	\includegraphics[width=0.13\textwidth]{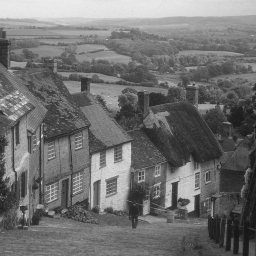}
	\includegraphics[width=0.13\textwidth]{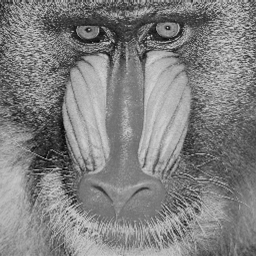}
    \includegraphics[width=0.13\textwidth]{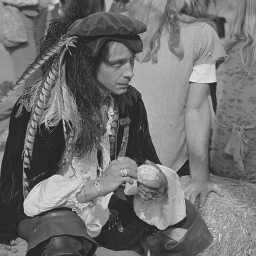}
	\caption{Test images. From left to right: Peppers, Bridge, Boat, Couple, Hill, Mandril, Pirate.}
	\label{figure_test_images}
\end{figure}

The standard test images of size $256 \times 256$ shown in Fig. \ref{figure_test_images} were utilized in the experiments.
Since the variance equals the expected value for Poisson noise, the peak intensity of the original image defines the noise level in the noisy image.
On one hand, it was shown in \cite{giryes2014sparsity} that the strategy referred to as binning is preferred when the peak intensity is less than 1.
According to this approach, the input image is first down-sampled by aggregating neighboring pixels,
which results in a smaller image with a higher peak intensity.
This image is then processed by any Poisson denoising algorithm and upscaled to the original size using interpolation.
On the other hand, when the peak intensity is higher than 5, the methods based on the variance stabilization are preferable.
Therefore, we varied the peak intensity between 1 and 5 in our experiments.

We compared our approach with the SPDA \cite{giryes2014sparsity} and the BM3D with the exact unbiased inverse Anscombe transformation \cite{makitalo2011optimal},
which are currently the best existing techniques.
Their code is available online and was taken with the default parameter settings.
The C++ implementation of our algorithm, which was utilized in the experiments, is available at http://physics.medma.uni-heidelberg.de/cms/content/poisson-denoising.
All overlapping patches from the BSD500 image set \cite{MartinFTM01} were taken as $\vu^{(1)}, \ldots, \vu^{(N_P)}$, which resulted in a set of $N_P \approx 7 \cdot 10^7$ patches.

\subsection{Choice of Parameters}

Increase of the patch size $\sqrt{d} \times \sqrt{d}$ makes it possible to handle more complex image structures
but requires a larger number of patches $N_P$ to represent the natural image patch distribution.
In our experiments, which were conducted with $N_P \approx 7 \cdot 10^7$, we used patches of size $14 \times 14$.

Regarding the number of clusters $N_C$, the larger it is, the more accurately estimator (\ref{MMSE_estimator_cluster}) approximates (\ref{MMSE_estimator_app}).
Nevertheless, $N_C$ cannot be arbitrary large since it should be computationally feasible to construct the k-d trees and the $K$-nearest neighbor graph.
Due to this reason, we set $N_C$ to $10^6$. % in our implementation.

Increase of the neighbor list size $K$ results in a larger number of clusters, which should be processed at each iteration of the loop in lines 10--14 of Algorithm 1.
On the other hand, it reduces the number of loop iterations, which are required to achieve sufficient denoising accuracy,
and makes it possible to set smaller $M$ in the convergence criterion.
We utilized $K = 2d$ and $M = 10$ in our experiments.

The number of k-d trees $N_T$ and the k-d tree leaf size $L$ do not significantly affect the results and were set to $N_T = 64$ and $L = 32$.

\subsection{Results}

The results are presented in Table \ref{table_results}.
For small peak intensities, the presented method outperforms the existing techniques by about 0.3 dB.
As the peak intensity increases, the results of our algorithm become comparable to those of \cite{makitalo2011optimal}.

% 0.28
% 0.29
% 0.15
% 0.04
% -0.07

\begin{table*}[h]
\caption{The denoising results. Each value is the average of five noise realizations. The best result is selected with the bold font. The PSNR is given in decibels.}
\label{table_results}
\begin{center}
\begin{tabular}{|c|c|ccccccc|c|}
\hline
Method                          & Peak  & \multicolumn{7}{c|}{Image}                                                                                           & Average        \\ \cline{3-9}
                                &       & Peppers        & Bridge         & Boat           & Couple         & Hill           & Mandril        & Pirate         &                \\ \hline

SPDA \cite{giryes2014sparsity}  &       & 19.96          & 19.21          & 20.07          & 20.05          & 20.61          & 18.40          & 20.06          & 19.77          \\
BM3D \cite{makitalo2011optimal} & 1     & 19.86          & 19.22          & 20.01          & 19.93          & 20.69          & 18.14          & 20.12          & 19.71          \\
proposed                        &       & \textbf{20.38} & \textbf{19.55} & \textbf{20.24} & \textbf{20.26} & \textbf{20.98} & \textbf{18.43} & \textbf{20.49} & \textbf{20.05} \\ \hline

SPDA \cite{giryes2014sparsity}  &       & 21.20          & 20.14          & 21.01          & 20.97          & 21.81          & 18.92          & 21.22          & 20.75          \\
BM3D \cite{makitalo2011optimal} & 2     & 21.94          & 20.31          & 21.07          & 21.00          & 21.63          & 18.79          & 21.28          & 20.86          \\
proposed                        &       & \textbf{22.26} & \textbf{20.65} & \textbf{21.28} & \textbf{21.22} & \textbf{22.05} & \textbf{18.98} & \textbf{21.60} & \textbf{21.15} \\ \hline

SPDA \cite{giryes2014sparsity}  &       & 21.91          & 20.43          & 21.34          & 21.22          & 22.13          & 19.05          & 21.60          & 21.10          \\
BM3D \cite{makitalo2011optimal} & 3     & 23.35          & 21.00          & 21.86          & 21.65          & 22.48          & 19.28          & 22.04          & 21.67          \\
proposed                        &       & \textbf{23.37} & \textbf{21.21} & \textbf{21.93} & \textbf{21.81} & \textbf{22.78} & \textbf{19.35} & \textbf{22.29} & \textbf{21.82} \\ \hline

SPDA \cite{giryes2014sparsity}  &       & 22.06          & 20.55          & 21.51          & 21.34          & 22.16          & 19.11          & 21.86          & 21.23          \\
BM3D \cite{makitalo2011optimal} & 4     & \textbf{24.03} & 21.50          & \textbf{22.36} & \textbf{22.26} & 23.06          & \textbf{19.58} & 22.61          & 22.20          \\
proposed                        &       & 23.92          & \textbf{21.60} & 22.32          & \textbf{22.26} & \textbf{23.23} & 19.56          & \textbf{22.79} & \textbf{22.24} \\ \hline

SPDA \cite{giryes2014sparsity}  &       & 22.31          & 20.60          & 21.60          & 21.45          & 22.22          & 19.16          & 22.04          & 21.34          \\
BM3D \cite{makitalo2011optimal} & 5     & \textbf{24.67} & 21.87          & \textbf{22.84} & \textbf{22.65} & 23.51          & \textbf{19.82} & 23.03          & \textbf{22.63} \\
proposed                        &       & 24.40          & \textbf{21.90} & 22.68          & 22.55          & \textbf{23.61} & 19.72          & \textbf{23.09} & 22.56          \\ \hline

\end{tabular}
\end{center}
\end{table*}

It took, on average, 14 minutes for our method implemented in C++ to process one 256 $\times$ 256 image on a PC with CPU Intel i7-5820K 3.3 GHz.

\section{Discussion and Conclusion}

In this work, we present a new noise suppression method and demonstrate its applicability to Poisson noise removal.
Compared with the approaches \cite{giryes2014sparsity,salmon2014poisson,makitalo2011optimal},
our algorithm has a different assumption about the original image.
Instead of utilizing image self-similarity or sparse patch representation,
patches of the original image are assumed to be sampled from the distribution representing natural image patches.
As a result, denoised patches are computed using the MMSE estimator, i.e. as the conditional expectation of the original patch given the noisy patch (\ref{cond_exp}).
As pointed out in \cite{levin2012patch}, such approach would provide optimal denoising results if the number of patches $N_P$ tended to infinity and the patch size $d$ was increased to the image size.
Nevertheless, it is useful also with fixed $N_P$ and $d$.
As the experiments showed, it is the preferable method for peak intensities below 4, for which the variance stabilization is not efficient enough.

Since a large number of patches is necessary to represent the image patch distribution,
the choice of suitable data structures is of primary importance in order to calculate the MMSE estimator with a reasonable computational effort and make it practically applicable.
In this work, it is demonstrated that
the k-d tree search followed by the search in the $K$-nearest neighbor graph allows fast and accurate calculation of the MMSE estimator.

\bibliographystyle{plain}
\bibliography{references}

\end{document}